\documentclass[11pt]{article}
\usepackage{ifthen}
\usepackage{mdwlist}
\usepackage{amsmath,amssymb,amsfonts,amsthm}
\usepackage{bm}
\usepackage[colorlinks=true,linkcolor=blue,citecolor=blue,urlcolor=blue]{hyperref}
\usepackage{enumitem}
\usepackage{graphicx}
\usepackage{xspace}
\usepackage{verbatim}
\usepackage{algorithm}
\usepackage{algpseudocode}
\usepackage[margin=1in]{geometry}
\usepackage{color}
\usepackage{thm-restate}
\usepackage{latexsym}
\usepackage{epsfig}
\usepackage[colorinlistoftodos,textsize=scriptsize]{todonotes}
\usepackage{booktabs}  
\usepackage{pifont}  
\usepackage{soul}  

\def\withcolors{1}
\def\withnotes{1}
\ifnum\withcolors=1
  \newcommand{\gcolor}[1]{{\color{green}#1}}
  \newcommand{\pcolor}[1]{{\color{red}#1}}
  \newcommand{\ncolor}[1]{{\color{orange}#1}}
\else
  \newcommand{\gcolor}[1]{{#1}}
  \newcommand{\pcolor}[1]{{#1}}
  \newcommand{\ncolor}[1]{{#1}}
\fi

\ifnum\withnotes=1
  \newcommand{\gnote}[1]{\par\gcolor{\textbf{G: }\sf #1}} 
  \newcommand{\gfootnote}[1]{\footnote{{\bf \gcolor{Gautam}}: {#1}}}
  \newcommand{\pnote}[1]{\par\gcolor{\textbf{P: }\sf #1}} 
  \newcommand{\pfootnote}[1]{\footnote{{\bf \pcolor{Pranav}}: {#1}}}

  \newcommand{\pranav}[1]{\par\pcolor{\textbf{P: }\sf #1}}

  \newcommand{\nic}[1]{\ncolor{\sf #1}}
  \newcommand{\nicx}[1]{\ncolor{\st{#1}}}
  \newcommand{\pranavx}[1]{\pcolor{\st{#1}}}

\else
  \newcommand{\gnote}[1]{}
  \newcommand{\gfootnote}[1]{}
  \newcommand{\pnote}[1]{}
  \newcommand{\pfootnote}[1]{}
  \newcommand{\pranav}[1]{}
  \newcommand{\nic}[1]{}
  \newcommand{\nicx}[1]{}
  \newcommand{\pranavx}[1]{}

\fi
\newcommand{\ignore}[1]{\leavevmode\unskip} 

\newcommand{\ctfp}{Custom TFP\xspace}
\newcommand{\vmap}{\texttt{VMAP}\xspace}
\newcommand{\jit}{\texttt{JIT}\xspace}

\newcommand{\anon}[1]{#1}

\title{Enabling Fast Differentially Private SGD \\ via Just-in-Time Compilation and Vectorization\thanks{Authors PS and NV have equal contribution.}}

\author {
  \anon{
  Pranav Subramani\thanks{Cheriton School of Computer Science, University of Waterloo. {\tt pranav.subramani@uwaterloo.ca}.}
  \and
  Nicholas Vadivelu\thanks{Cheriton School of Computer Science, University of Waterloo. {\tt nbvadive@uwaterloo.ca}.}
  \and
  Gautam Kamath\thanks{Cheriton School of Computer Science, University of Waterloo. {\tt g@csail.mit.edu}. Supported by an NSERC Discovery grant, a Compute Canada RRG grant, and a University of Waterloo startup grant.}
  }
}

\begin{document}
\maketitle

\begin{abstract}
  A common pain point in differentially private machine learning is the significant runtime overhead incurred when executing Differentially Private Stochastic Gradient Descent (DPSGD), which may be as large as two orders of magnitude.
  We thoroughly demonstrate that by exploiting powerful language primitives, including vectorization, just-in-time compilation, and static graph optimization, one can dramatically reduce these overheads, in many cases nearly matching the best non-private running times.
  These gains are realized in two frameworks: one is JAX, which provides rich support for these primitives through the XLA compiler.
  We also rebuild core parts of TensorFlow Privacy, integrating more effective vectorization as well as XLA compilation, granting significant memory and runtime improvements over previous release versions.
  Our proposed approaches allow us to achieve up to 50x speedups compared to the best alternatives.
  \anon{Our code is available at \url{https://github.com/TheSalon/fast-dpsgd}.}
\end{abstract}

\section{Introduction}

Machine learning has recently experienced tremedous growth, being used to solve problems with unprecedented accuracy in a myriad of domains.
However, not all domains are alike.
One may wish to train a model on a dataset which is publicly available---for instance, training an digit classifier on the popular (public) MNIST database~\cite{LeCunBBH98}.
On the other hand, applications frequently involve datasets which are in some way \emph{private}, potentially containing data which is personal information, or otherwise proprietary.
To provide some examples, consider a medical application, where one wishes to classify whether or not images contain a tumor.
Alternatively, one can imagine a retail company training a machine learning model based on valuable market research data.
In both cases, it is of paramount importance that the trained model does not leak information about the training data, with consequences ranging from loss of revenue to legal liability.
Troublingly, it has been demonstrated that disregarding these concerns can result in significant leakage of private information---for example, a language model na\"ively trained on a sensitive dataset may end up regurgitating Social Security numbers~\cite{CarliniLEKS19}.
Furthermore, many heuristic and best-effort privacy approaches (such as data anonymization) have been demonstrated to be non-private~\cite{SamaratiS98, BhowmickDFKR18}.

In order to assuage concerns of private information leakage, in 2006, Dwork, McSherry, Nissim, and Smith~\cite{DworkMNS06} introduced the celebrated notion of differential privacy (DP).
This principled and rigorous measure is widely accepted as a strong standard for privacy-preserving data analysis.
Informally speaking, an algorithm is said to be differentially private if its distribution over outputs is insensitive to the addition or removal of a single datapoint from the dataset.
Differential privacy has enjoyed widespread adoption in practice, including deployments by Apple~\cite{AppleDP17}, Google~\cite{ErlingssonPK14,BittauEMMRLRKTS17}, Microsoft~\cite{DingKY17}, and the US Census Bureau for the 2020 Census~\cite{DajaniLSKRMGDGKKLSSVA17}.

One of the workhorse algorithms in machine learning is stochastic gradient descent (SGD), which is effective at training machine learning models in rather general settings.
A differentially private analogue, DPSGD~\cite{SongCS13, BassilyST14, AbadiCGMMTZ16}, has been introduced as a drop-in replacement for SGD.
This algorithm can (informally) be described as iteration of the following simple procedure:
\begin{enumerate}
  \item Draw a minibatch from the training dataset of appropriate size;
  \item For each point $(x_i,y_i)$ in the minibatch:
    \begin{enumerate}
      \item Compute the gradient $g_i$ of the objective function at $(x_i,y_i)$;
      \item ``Clip'' the gradient: if $\|g_i\|_2$ is greater than some hyperparameter threshold $C$, rescale $g_i$ so that $\|g_i\|_2 = C$;
    \end{enumerate}
  \item Aggregate the clipped gradients in the minibatch, and add Gaussian noise of sufficient magnitude to guarantee differential privacy;
  \item Update the parameters of the model by taking a step in the direction of the noised aggregated gradient.
\end{enumerate}
The primary differences with respect to (non-private) SGD are the per-example clipping and noising operations.
While these modifications seem relatively innocuous, they have so far led to non-trivial costs in terms of both running time as well as accuracy of the trained model.
In this paper, we address and mitigate the running time overhead of DPSGD.\footnote{This article concentrates on DPSGD, as it is the most commonly run private algorithm on large-scale datasets. However, we note that this per-example clipping operation is present in numerous differentially private algorithms~\cite{KarwaV18, KamathLSU19, KamathSU20, BiswasDKU20}, and similar methods might be useful in performance optimization in these settings as well.}

Taking a step back: as we alluded to before, the deep learning revolution has been catalyzed by advances in graphics processing units (GPUs) and similar devices which allow simultaneous processing of several datapoints at once.
More precisely, they allow one to compute gradients for each example, which are then aggregated.
This specific procedure is ubiquitous in machine learning, and has thus been highly optimized in most machine learning frameworks.

In fact, one could even consider this procedure to be \emph{too} optimized: in vanilla machine learning settings, one simply needs the gradient as averaged over a minibatch, and not for each individual point.
Consequently, modern machine learning frameworks generally do not allow access to gradients for individual points, also known as \emph{per-example gradients}.
Access to these objects is critical for the clipping procedure in DPSGD, as well as other applications of independent interest beyond privacy, for instance optimization based on importance sampling~\cite{ZhaoZ15}.
If one wishes to generate per-example gradients, the immediate solution is to process the gradients one by one, thus losing all advantage bestowed by parallel processing on GPUs and creating massive overheads in terms of the running time.
Lack of support for fast computation of per-example gradients has been noted and lamented numerous times for both TensorFlow~\cite{Lipton16, seerdecker16, act6517} and PyTorch~\cite{Kunstner18, alexdepremia18}.
PyTorch co-creators, Chintala and Paszke, have both commented on this issue, stating in 2018 that ``this is currently impossible with [auto differentiation] techniques we use''~\cite{Paszke18}, due to limitations with the THNN and cuDNN backends~\cite{Chintala17}, and adding support for this functionality would require ``chang[ing] ~5k+ lines of code''~\cite{Chintala18}.



Numerous attempts to avoid these computational roadblocks have been proposed.
Goodfellow~\cite{Goodfellow15} proposed an algorithmic solution for computing per-example $\ell_2$-norms of the gradients for fully-connected networks.\footnote{Though he was motivated by other applications~\cite{ZhaoZ15}, these are the exact objects we require for DPSGD.}
Other proposed solutions work by exploiting Jacobians~\cite{DangelKH20} or parallelizing over the batch dimension~\cite{AgarwalG19}.
Several of these approaches are are restricted to specific types of architectures---for example, \cite{Goodfellow15} is restricted to fully-connected layers, though~\cite{RochetteMT19} extends this to convolutional layers, and a very recent work~\cite{LeeK20} further considers layers including recurrent networks, attention, and more.
BackPACK~\cite{DangelKH20} currently supports only fully connected and convolutional layers, and while the paper states that it can be extended to recurrent and residual layers, GitHub issues related to implementation of these features have been open since November 2019~\cite{backpackrnn}.
Facebook's Opacus~\cite{YousefpourSSTPMNGBZCM21} emphasizes speed and scalability as the main selling points.
We briefly mention microbatching, in which comparatively small subsets of the minibatch called ``microbatches'' of points are averaged before clipping, reducing the number of clipping operations (and thus the running time), at the cost of requiring additional noise to achieve the same privacy guarantee.
Since this generally results in significantly worse accuracy, we do not investigate it further in our work.
A more thorough description of approaches is provided in Section~\ref{sec:approaches}.

As mentioned in the literature (and thoroughly explored in this paper), all existing approaches seem to incur moderate to severe running time overhead versus non-private SGD, with slowdowns as large as two orders of magnitude.
For instance, Carlini et al.~\cite{CarliniLEKS19} comment, ``Training a differentially private algorithm is known to be slower than standard training; our implementation of this algorithm is 10-100x slower than standard training,'' where their implementation is based on TensorFlow Privacy.
Additionally, Thomas et al.~\cite{ThomasADMK20} (working in an unspecified framework) document a slowdown from 12 minutes to 14 hours due to the introduction of differential privacy, a 70x slowdown.
The effect of these slowdowns can range from an inconvenience when it comes to rapid prototyping of smaller models, to prohibitively expensive for a single training run of a larger model.
Overcoming this obstacle is an important step in helping differentially private machine learning transition from its present nascent state to widespread adoption.

\subsection{Results}
We demonstrate that one can mostly eliminate the significant running time overhead of differentially private SGD by exploiting language primitives such as vectorization, just-in-time (JIT) compilation, and static graph optimization.
These features are core primitives within JAX~\cite{FrostigJL18, BradburyFHJLMW18} and TensorFlow 2 (TF2)~\cite{tensorflow2015}, both tensor-processing libraries from Google.
These frameworks combine JIT compilation backed by the Accelerated Linear Algebra (XLA)~\cite{XLA} just-in-time compiler (JIT) with auto differentiation for high-performance machine learning.
As we will see, JAX is consistently the fastest method for running DPSGD, with running times comparable to the non-private case.
Our custom TensorFlow Privacy (TFP) implementation (referred to as \ctfp), which leverages vectorization and XLA compilation in TensorFlow 2, demonstrates similar performance to JAX and significantly outperforms the existing TFP library.
These changes have since been merged into TensorFlow Privacy.


Our primary contributions are as follows:
\begin{enumerate}
  \item We thoroughly benchmark several frameworks and libraries for DPSGD.
  \item We extend TensorFlow Privacy to support TF2, more efficient vectorization, and XLA compilation, significantly improving its running time in most cases (referred to as \ctfp in this paper).
  We also contribute a variant of our implementation to TensorFlow Privacy, which is now the fastest DPSGD algorithm the library provides.
  \item We demonstrate that methods which use vectorization, JIT compilation, and static graph optimization are consistently the fastest and most memory-efficient: specifically, JAX and \ctfp.
  \item We find that, despite similarities in the compilation pipeline, JAX is generally faster than \ctfp. We examine and discuss compiled XLA assembly to explain the discrepancy.
  \item We provide developers of DP machine learning libraries and frameworks with recommendations to optimize running time performance.
  \item Finally, we publicly release code to reproduce these experiments, as well as guide researchers and engineers in producing fast code for differentially private machine learning.
    \anon{Code is available at \url{https://github.com/TheSalon/fast-dpsgd}.}
\end{enumerate}

Table~\ref{tab:headline} summarizes some of our experimental results, with median running time per epoch for a variety of settings.



\begin{table*}[h!]
  \centering
  \begin{tabular}{lrrr}
    \toprule
     Architecture         & Private JAX/\ctfp   & Best Private Alternative & Best Non-Private \\
    \midrule
    Logistic Regression  & {\bf 0.23}  & 0.48 & 0.12 \\
    FCNN           & {\bf 0.21}  &  0.77 & 0.20 \\
    MNIST CNN      & {\bf 0.53}  &  6.50 & 0.42 \\
    CIFAR10 CNN    & {\bf 7.3}  &  12 & 1.7  \\
    Embedding      & {\bf 0.23}  &  3.8  & 0.11  \\
    LSTM           & {\bf 8.2}   &  407  & 3.6  \\
    \bottomrule
  \end{tabular}
  \caption{Median running time (s) per epoch of training various models at batch size 128, both with and without differential privacy, comparing our suggested solutions (JAX and \ctfp) with other frameworks. FCNN stands for Fully-Connected Neural Network, CNN stands for Convolutional Neural Network, and LSTM stands for Long-Short Term Memory network. JAX or \ctfp are consistently the fastest private options with little overhead over the best non-private variants.
  }
  \label{tab:headline}
\end{table*}

We observe dramatic improvements for embedding network and LSTMs~\cite{hochreiter1997long}, for the latter, potentially significant enough to bring LSTMs from impractical into the realm of feasibility.
JAX is able to privately train these models $17$x and $50$x faster than the best alternative.\footnote{Note that a confirmed bug in TF2 currently prevents us from running \ctfp in these cases~\cite{embedxla}.}
Examining the overhead due to privacy: JAX's running time increases by roughly $2$x, compared to factors closer to $10$x for alternatives.
We take this as evidence that optimization and improvements to the core JAX framework (which is still relatively young) will translate to further advantages for private training.

For the fully-connected and MNIST convolutional networks, JAX or \ctfp almost entirely remove the overhead due to privacy.
In fact, the running times are significantly better than some alternatives \emph{without} privacy.
Recall that these are per-epoch times: while an improvement of $0.5$ seconds might seem insignificant, this can add up when training for many epochs.
We perform an ablation study (Tables~\ref{tab:jaxablation} and~\ref{tab:tfablation}) for some models to pinpoint the source of all improvements.

While there has been significant work towards making DPSGD run faster, they are without exception built \emph{on top} of frameworks such as PyTorch and TensorFlow.
By investigating the effect of low-level language features, we thus provide a qualitatively different (and more systems-focused) perspective on the problem, which we hope will refocus efforts by the community to speed up private machine learning.
We discuss how other frameworks can take advantage of this perspective in Section~\ref{sec:recs}.
Briefly, we recommend improved support for vectorization and JIT compilation.

While our investigations show the consistent and substantial superiority of JAX for fast private machine learning, these benefits remain relatively unknown.
Though a small number of experts are aware~\cite{Talwar20}, and the official JAX repo contains a toy demonstration~\cite{Creager19}, before the initial posting of our paper a Google Scholar search revealed only two papers which use JAX for differential privacy~\cite{WaitesC20, papernot2020tempered}, and neither emphasizes or even comments on the computational advantages of JAX.
Similarly, while efficient per-example gradients have been studied in TensorFlow~\cite{agarwal2019autovectorizing}, efficient application of these techniques is not readily available to privacy researchers.
We hope that our investigation will document this phenomenon and encourage others to adopt it for their private machine learning needs.


\subsection{Simultaneous and Subsequent Work}
Simultaneous to our work, \cite{BuGKLST21} employ Johnson-Lindenstrauss projections to quickly approximate per-example gradient norms.
This is an algorithmic modification, and will not be functionally equivalent to DPSGD -- similar to microbatching, there is a time-accuracy tradeoff (though not as severe in this case).

Subsequent to the initial posting of this paper, we worked with Google engineers to implement our improvements into TensorFlow Privacy.
Vadivelu contributed an JAX implementation of DPSGD to the Optax library~\cite{Optax20}.
\cite{AnilGGKM21} employed our findings to efficiently privately train BERT-Large.
In a recent Opacus whitepaper~\cite{YousefpourSSTPMNGBZCM21}, the authors repeat some of our experiments on more recent versions of these frameworks; we defer to their work for discussion of these results.

\section{Description of Approaches}
\subsection{Libraries Enabling DPSGD}

\paragraph{JAX~\cite{FrostigJL18, BradburyFHJLMW18}.}
JAX is a recently introduced framework for machine learning, defined by its automatic differentiation capabilities and JIT compilation via the XLA compiler \cite{XLA}.
Programs written in pure Python and JAX's NumPy~\cite{harris2020array} API can be translated to an intermediate language (XLA-HLO) to be JIT compiled, i.e., to generate custom assembly instructions for the hardware.
This enables optimizations such as kernel-fusions, buffer reuse, improved memory layout, and more.
Additionally, one of the core functions present in JAX is \vmap, a vectorized map, which enables easy-to-write and efficient batch level parallelism that is fundamental to DPSGD.
As we will demonstrate, these enable the fastest approach for DPSGD that we are aware of.

\paragraph{\ctfp.}
Vectorization and XLA-driven JIT compilation is also available in TensorFlow 2~\cite{tensorflow2015}, which we leverage in our implementation, \ctfp.
With these primitives, we achieve performance comparable to JAX and surpassing existing DPSGD implementations in TensorFlow.  
We augment TensorFlow Privacy to better utilize \texttt{tf.vectorized\_map} and follow TensorFlow 2 best practices while retaining the existing functionality. 

\paragraph{Chain-Rule-Based Per-Example Gradients~\cite{Goodfellow15, RochetteMT19}.}
This suite of techniques is implemented on top of PyTorch~\cite{pytorchNEURIPS2019}.
They support efficient GPU-accelerated per-example gradients for fully-connected layers via~\cite{Goodfellow15}, as well as convolutional layers via~\cite{RochetteMT19}, which we describe in the detail in the following paragraphs.

Let $C, D, T,$ and $B$ refer to the number of input channels, output channels, the spatial dimension, and the batch size.
The shape of the input $x$ is $(B, C, T)$.
The conventional formula for the discrete convolution can be written as: $$\sum_{c = 0}^{C - 1} \sum_{k = 0}^{K - 1} x[b,c,t+K]h[d,c,k].$$
 The gradient of this expression can be efficiently computed via automatic differentiation~\cite{rall1981automatic}.
 PyTorch's automatic differentiation cannot be parallelized across the batch dimension $b$~\cite{RochetteMT19}, which is required to backpropagate through the above expression.
 Instead, they rewrite the convolution as follows: $$\sum_{c = 0}^{C/G - 1} \sum_{k = 0}^{K - 1}x\left[b,c,g \frac{C}{G}, t+K\right] h[d,g,c,k],$$
  where $G$ is the number of groups and the shape of $x$ is $(1, B, C, T)$.
  The initial convolution is $1$-dimensional, while the above expression includes an added dimension.
  Similarly, to allow backpropagation through a $k$-dimensional convolutional layer, a $(k + 1)$-dimensional convolutional layer is required.
  This can be achieved by utilizing the \texttt{group} attribute in the convolution function in PyTorch, since splitting it into groups implies that the same convolution is applied to each individual group.

  \paragraph{BackPACK~\cite{DangelKH20}.}
  The chain rule gives the following expression for the gradient of a loss function:


  \[
    \nabla_{\theta^{(i)}} \ell(\theta) = (J_{\theta^{(i)}} z_n^{(i)})^T \left(\prod_{j=i}^{L-1} (J_{z_n^{(j)}} z_n^{(j + 1)})^T \right) (\nabla_{z_n^{(L)}} \ell_{n} (\theta)).
  \]
  In order to compute this quantity, one requires the ability to multiply the Jacobian by a vector and by a matrix, which is not currently supported in PyTorch's automatic differentiation framework.
  In BackPACK, Dangel et al.~\cite{DangelKH20} extend several layers within PyTorch to support fast Jacobian-vector and Jacobian-matrix products in order to extract quantities like individual gradients, variance, $\ell_2$-norm of the gradients, and second-order quantities.
  In particular, to extract first-order gradients, their method multiplies the transposed Jacobian with the outputs of the layer:
  $$\frac{1}{N}\nabla_{\theta^{(i)}} \ell(\theta) = \frac{1}{N} (J_{\theta^{(i)}} z_n^{(i)})^T (\nabla_{z_n^{(i)}} \ell(\theta)),$$
  where $i = 1, \dots, N$ and each $\theta^{(i)}$ has a gradient which is of shape $(N, d^{(i)})$.
  BackPACK provides efficient computation for the transpose of the Jacobian as well as the Jacobian.
  This method compares favourably to extracting gradients sequentially via a for-loop.
  The authors mention that the cost of extracting the individual gradients does come at a minor overhead when compared to regular training.\footnote{By regular training we mean evaluating the gradients on the average loss for the batch.}

  \paragraph{Opacus~\cite{YousefpourSSTPMNGBZCM21}.}
  Opacus is a library for training PyTorch models with differential privacy, recently released by Facebook.
  It supports per-example gradients, using PyTorch's forward and backward hooks to propagate gradients.
  They provide support for several PyTorch layers including LSTM layers, which are not supported in either of the previous two frameworks.
  Note that Opacus does not support PyTorch's \texttt{nn.LSTM} but instead implements a separate \texttt{opacus.layers.DPLSTM}, with adjustments that allow individual gradients to propagate through it.

  \paragraph{PyVacy~\cite{pyvacy19}.}
  Before the release of Opacus (and its predecessor PyTorch-DP), PyVacy was the most popular library for DP machine learning in PyTorch.
  PyVacy has no custom support for parallelization across the batch dimension for any layer since it processes each sample individually (by way of a for-loop).
  This generally leads to a large increase in runtime for models trained using PyVacy.

  \paragraph{TensorFlow Privacy~\cite{tfp18}}  TensorFlow Privacy is a library for differentially private machine learning, built on top of TensorFlow. 
  TensorFlow Privacy has general support for a vectorized implementation of DPSGD via \texttt{vectorized\_map} which allows it to parallelize across the batch dimension, used to extract per-example gradients.
  The library recently introduced a TensorFlow 2 compatible API that leverages \texttt{GradientTape.jacobian} to compute per-example gradients, which we compare seperately to the TensorFlow 1 API in our experiments.

\label{sec:approaches}

\subsection{Notable Framework Features} 

\paragraph{Static versus Dynamic Graph.} TensorFlow and JAX use a \emph{static graph} to track computation in order to optimize execution and compute gradients.
This means the sequence of operations is traced and a large proportion of shapes are determined during the first invocation of the function, allowing for kernel fusion, buffer reuse, and other optimizations on subsequent calls.
PyTorch uses a \emph{dynamic graph} to track computation flow in order to compute gradients, but does \emph{not} optimize execution.
This enables increased dynamism in the shapes and types of computations, at the cost of losing all the aforementioned optimizations.

\paragraph{Grappler versus XLA.} TensorFlow has two optimization engines: Grappler~\cite{GRAPPLER} and XLA.
Grappler, TensorFlow's original graph optimizer, takes as input the computation graph and is able to prune dead nodes, remove redundant computation, improve memory layouts, and more.
XLA, TensorFlow's new optimizing just-in-time compiler, can perform the same optimizations as Grappler, in addition to generating code for fused kernels.
For this reason, XLA has the potential to extract more performance out of TensorFlow graphs than Grappler, but does not always accomplish this due to Grappler's maturity.

\paragraph{JAX and XLA.} JAX was built from the ground up to leverage XLA, and so many of its operations map directly to XLA primitives.
We often observe that JAX is able to extract better performance out of XLA than TensorFlow.


\paragraph{Pytorch and Static Graphs.} Recently, PyTorch has released the capability to JIT compile its code through \texttt{torch.jit} or PyTorch XLA~\cite{torchxla}.
Due to the early nature of these two efforts, they were not successful in JIT compiling the methods we tried, thus we do not consider them further.

\paragraph{Just-In-Time Compilation} JIT compilation is a method of compilation that happens at runtime, as opposed to before program execution.
JAX and TensorFlow perform JIT compilation by recording operations that are executed on tensors/arrays (i.e. ``tracing''), generating the low-level instructions to perform these operations, optimizing these instructions, then producing fast low-level kernels.
The XLA compiler requires all array shapes to be known at trace-time so it can statically determine how much memory should be allocated for each operation.
While compilation can be relatively slow compared to execution (\textasciitilde 10x the time), you only pay this price once at the first training iteration, provided the input shapes do not change.

\section{Empirical Findings}
\label{sec:experiments}
In this section, we present our experimental findings, comparing and benchmarking various alternatives for DPSGD.
\anon{Our code is publicly available at \url{https://github.com/TheSalon/fast-dpsgd}.}

We evaluate the aforementioned implementations of DPSGD in runtime and memory consumption on four datasets:
\textbf{MNIST}~\cite{lecun1998mnist}, a handwritten digit recognition dataset with 60,000 training images of size 28$\times$28 each,
\textbf{CIFAR10}~\cite{Krizhevsky09learningmultiple}, a dataset of small colour images with 60,000 training examples of size 32$\times$32$\times$3 each,
\textbf{IMDb}~\cite{imdb2011}, a movie review sentiment classification dataset with 25,000 training examples padded to a sequence length of 256 each,
and \textbf{Adult} ~\cite{Dua:2019}, containing 45,220 examples with 104 features, which was preprocessed via methods from~\cite{IyengarNSTTW19}.

We perform our evaluations on six different architectures.
We start with the smallest dataset, Adult, training a 105-parameter logistic regression model and a 5,532-parameter fully-connected neural network (FCNN).
Next, we train an MNIST classifier, using a convolutional neural network architecture with 26,010 parameters which we refer to as MNIST CNN.
Then, we train a CIFAR10 convolutional neural network classifier architecture with 605,226 parameters used by Papernot et al.~\cite{papernot2020tempered}.
We refer to this network as CIFAR10 CNN.
For IMDb, we use an LSTM network and Embedding network with 1,081,002 and 160,098 parameters, respectively, demonstrating the method on a relatively large model..
Full descriptions of our architectures appear in Section~\ref{sec:arch}.
This selection covers the common data and architecture types at realistic sizes for differentially private learning.
In particular, we did not consider the exceptionally large models which are not prevalent in non-private machine learning.
This is because the noise introduced by DPSGD scales with the square root of the number of parameters, so such large models typically get trivial accuracy, and more modest-sized architectures are preferred.
These architectures were selected because they resemble the models used in the DPSGD tutorials released by TensorFlow Privacy and Opacus.
The fully connected network and the logistic regression model are a mainstay in classification and therefore were included.
The LSTM model with this architecture is undocumented to the best of our knowledge and is an example of a larger model.

We compare a number of different methods for fast DPSGD, including JAX~\cite{FrostigJL18, BradburyFHJLMW18}, BackPACK~\cite{DangelKH20}, the chain-rule based method (CRB)~\cite{Goodfellow15, RochetteMT19}, Opacus~\cite{YousefpourSSTPMNGBZCM21}, PyVacy~\cite{pyvacy19}, TensorFlow Privacy (TFP)~\cite{tfp18} and our modification of TFP, dubbed \ctfp.
For TensorFlow based frameworks, we evaluate performance both with and without XLA JIT compilation in both TF 1 and TF 2.
We refer to TensorFlow 2 and JAX as \textit{modern XLA-compiled libraries}, since the XLA integration in the TensorFlow 1 implementations is limited.
When possible, we base our implementations on the idiomatic examples provided by the libraries.

These architectures and datasets are evaluated in terms of runtime at batch sizes 16, 32, 64, 128, and 256. 
This showcases a comparison of runtimes across a variety of batch sizes to present a holistic picture of the running times, as well as demonstrating the impact of memory utilization on runtime.
Each experiment was run for 20 epochs and the median epoch run-time is reported.
The variance of these experiments was low enough that the errors bars are negligible (full results provided in the supplement).
Outside of the initial compilation time required for the static graph frameworks, the runtime showed little variance between epochs, resulting in narrow confidence intervals for the epoch runtime (data available upon request).
We preprocess all the data in advance and use an identical generator-based dataloader for all frameworks, to ensure consistency.
The data is stored in the framework-appropriate array/tensor in advance on the CPU, then is transferred to the GPU before the forward pass (as is common practice).

All experiments were run on Ubuntu 18.04 with an Intel Core i7-7800X CPU (3.50 GHz, 6 cores), NVIDIA GTX Titan V GPU (12GB VRAM), and 32GB of RAM.
Though we do not include these results in the paper, we verified that the relative performance of these libraries generalized to a consumer grade setup by running these experiments with an Intel i7-8850H (2.60GHz, 6 cores), NVIDIA GTX 1050 Ti Max-Q (4GB VRAM), and 16GB of RAM.



\begin{figure*}[ht!]
  \centering
  \includegraphics[width=16cm]{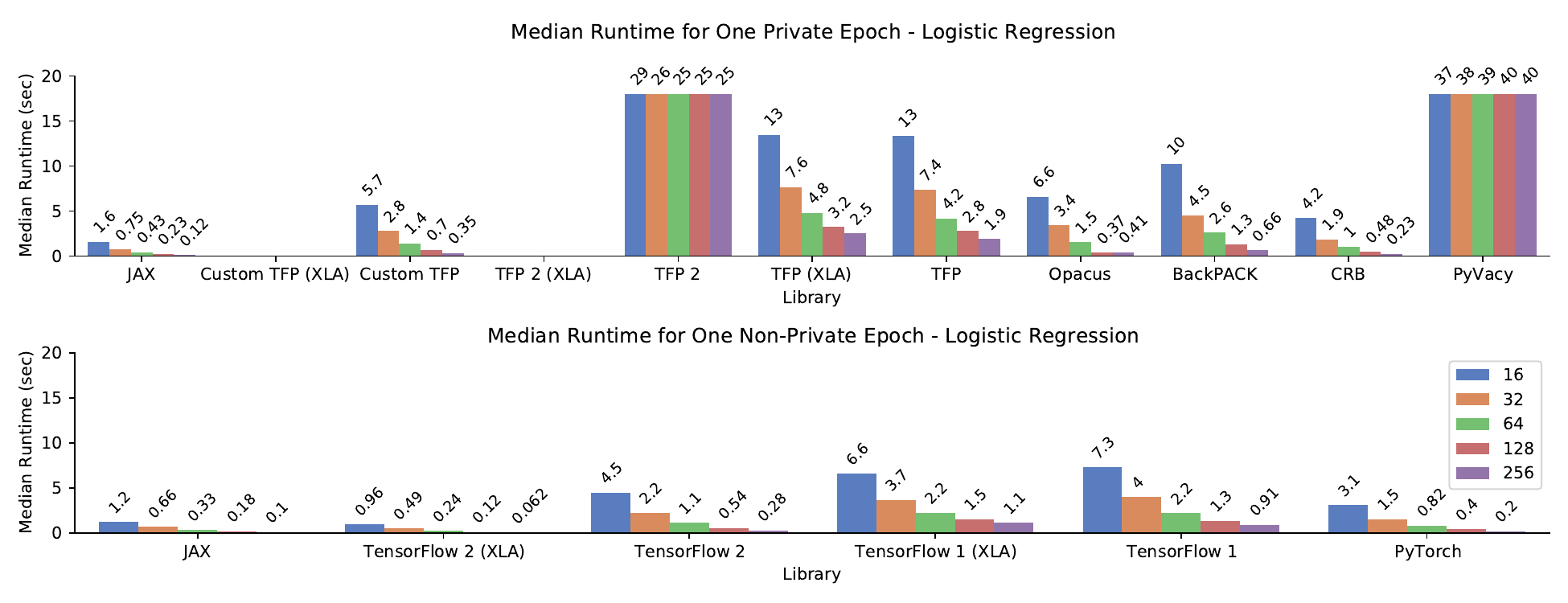}
  \caption{\textbf{Runtimes for logistic regression on the Adult dataset}.
    With privacy, JAX is the fastest, comparable to the non-private runtimes.
    We were unable to benchmark \ctfp due to an open TensorFlow 2 bug~\cite{logregxla}.
    The y-axis is truncated for clarity.}
  \label{fig:logregplot}
\end{figure*}

\begin{figure*}[ht!]
  \centering
  \includegraphics[width=16cm]{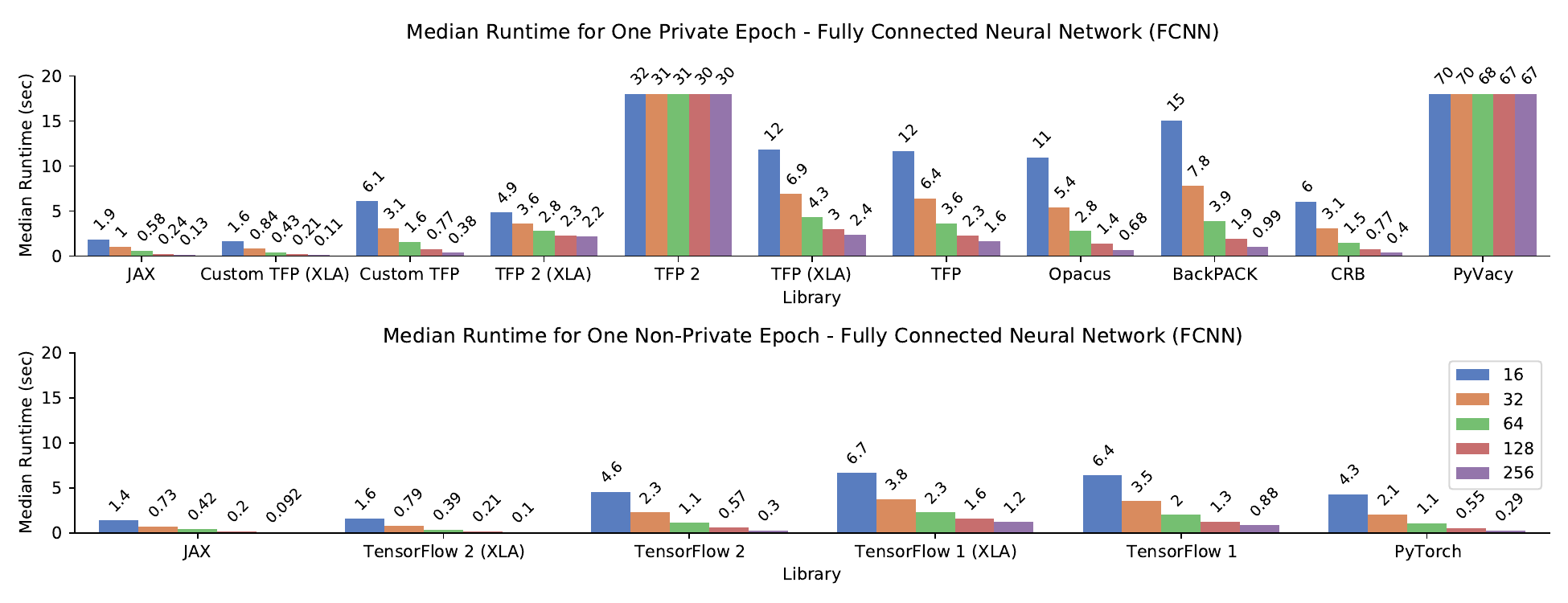}
  \caption{
    \textbf{Runtimes for the fully connected network on the Adult dataset}.
    We observe that JAX and \ctfp are the fastest by a large margin in both settings, with DPSGD having low overhead over the non-private setting.
    The y-axis is truncated for clarity.
   }
  \label{fig:ffnnplot}
\end{figure*}

Our first experiment (Figure~\ref{fig:logregplot}) is on the runtime performance of the logistic regression model.
We see that JAX has by far the most performant implementation of DPSGD.
\ctfp with XLA was unable to run due to a confirmed bug in TensorFlow 2~\cite{logregxla}.
However, considering TensorFlow 2 (XLA)'s performance in the non-private setting, we would expect similarly strong performance in the private setting once this is resolved.
TFP 2 without XLA performs unexpectedly poorly, due to the unoptimized jacobian computation which does not account for per-example independence of gradients.

\begin{figure*}[ht!]
  \centering
  \includegraphics[width=16cm]{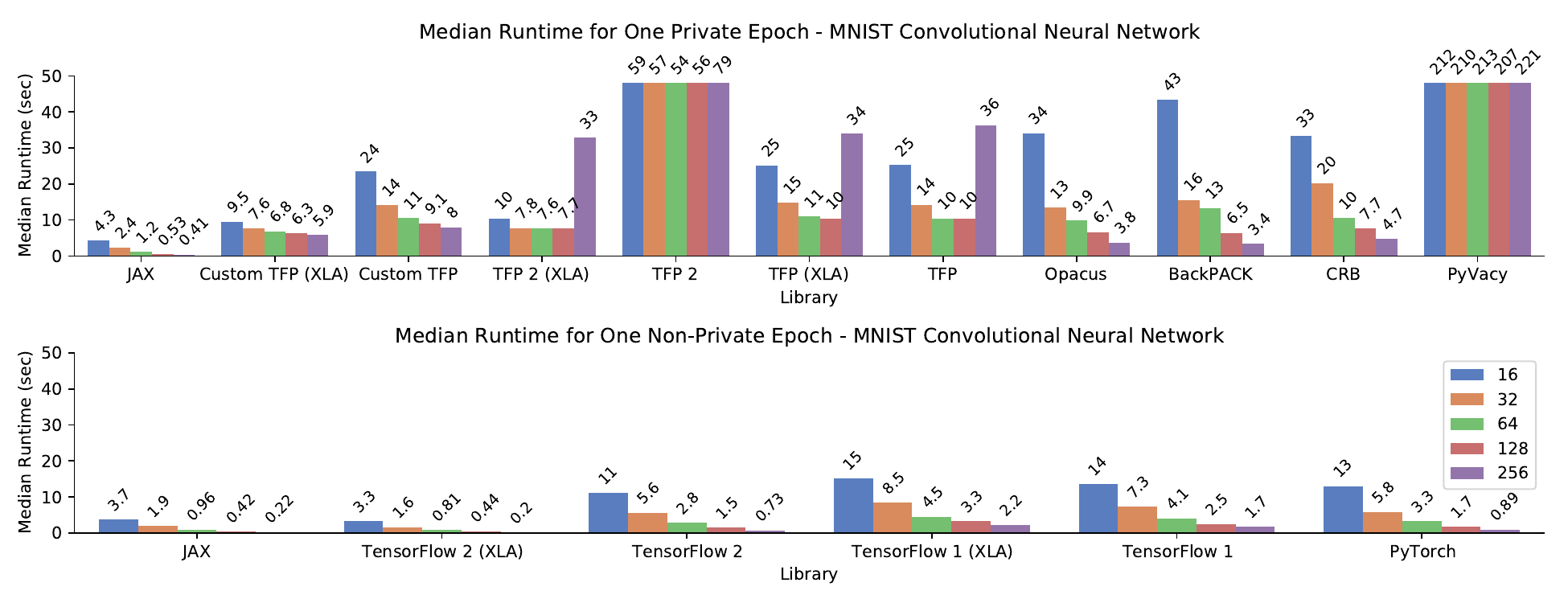}
  \caption{
    \textbf{Runtimes for the CNN on the MNIST dataset.}
    We observe that JAX has the fastest runtime in the private case while TensorFlow 2 with XLA is the fastest in the non-private case.
    TFP struggles at the largest batch size due to an inability to properly parallelize per-gradient computation with this level of memory consumption.
    The y-axis is truncated for clarity.
   }
  \label{fig:cnnplot}
\end{figure*}

\begin{figure*}[ht!]
  \centering
  \includegraphics[width=16cm]{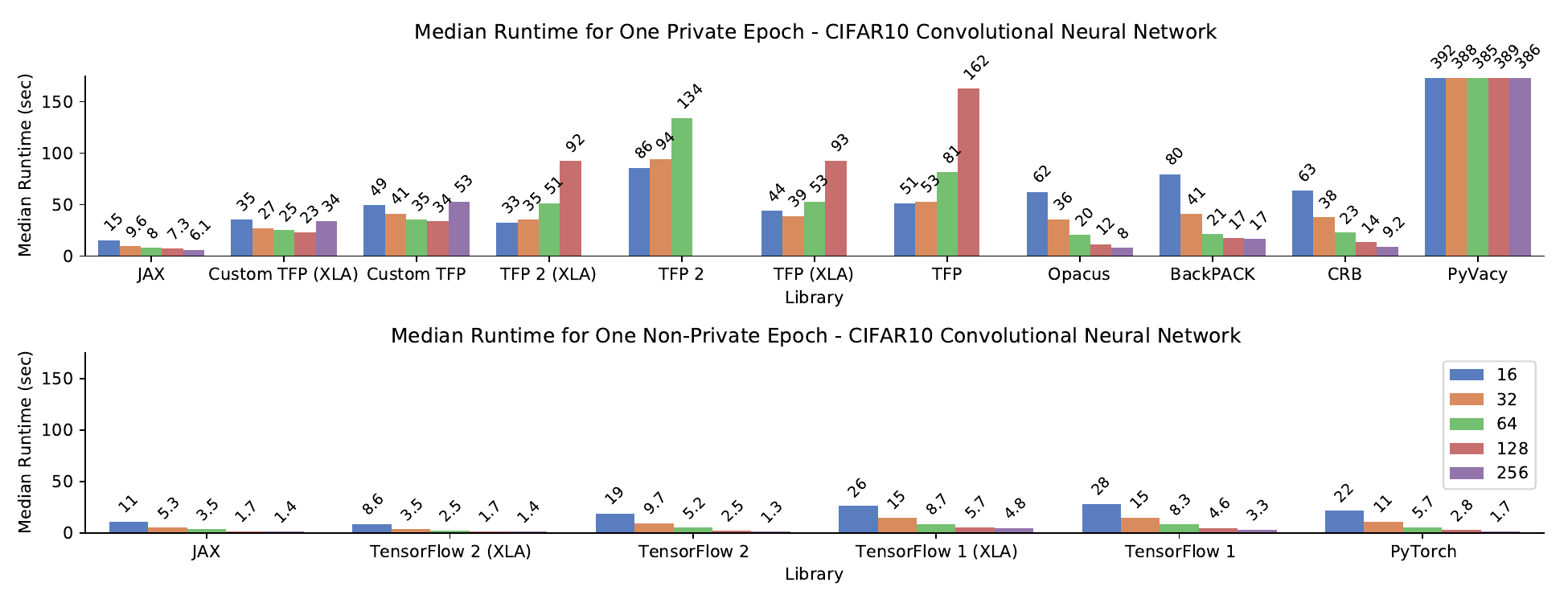}
  \caption{
    \textbf{Runtimes for the CNN on the CIFAR10 dataset.}
    We observe that JAX has the fastest runtime in the private case while TensorFlow 2 with XLA and JAX are the fastest in the non-private case at most batch sizes.
    Similar to the MNIST case, TFP struggles at the largest batch size due to an inability to properly parallelize per-gradient computation with this level of memory consumption.
    The y-axis is truncated for clarity.
   }
  \label{fig:cifar10cnnplot}
\end{figure*}

Next, we evaluate the FCNN model (Figure~\ref{fig:ffnnplot}), where we observe that the two modern XLA-compiled implementations (JAX and \ctfp) are significantly faster than the other options in both the private and non-private setting.
With such a simple architecture, the compiler can perform significant optimizations.
Notably, JAX and \ctfp show little overhead over their non-private counterparts.
The non-statically compiled frameworks (apart from PyVacy) remain competitive in this setting due to the shallow network size and low parameter count.
TFP 2 without XLA performs poorly again, but XLA optimization brings its performance to more reasonable levels.

We then evaluate the MNIST CNN model (Figure~\ref{fig:cnnplot}), where JAX is by far the most performant implementation in the private case, and the modern XLA-compiled frameworks are the fastest non-private methods.
\ctfp is noticeably slower than JAX---we conjecture that this is due to different utilization of the JIT compiler, see Section~\ref{sec:discussion}.
We see the PyTorch-based frameworks are much slower at small batch sizes: in this regime, the Python overhead is too high for the asynchronous execution to hide the latency.
Also, at the largest batch size, TFP's (including TFP 2) performance deteriorates due to the memory consumption, so it is no longer able to effectively vectorize the per-example gradient computation.

The CIFAR10 CNN model runtimes (Figure~\ref{fig:cifar10cnnplot}) demonstrate similar trends to the MNIST CNN.
In the TFP case, we observe a more exaggerated performance deterioration at larger batch sizes due to the larger model, and begin to see similar effects in \ctfp as well, showing performance worse than the PyTorch frameworks at the largest batch sizes.
The PyTorch libraries show closer performance to the compiled libraries at lower batch sizes unlike the MNIST case: since the model and data are larger, PyTorch's asynchronous execution is able to hide the Python latency.

\begin{figure*}[ht!]
  \centering
  \includegraphics[width=16cm]{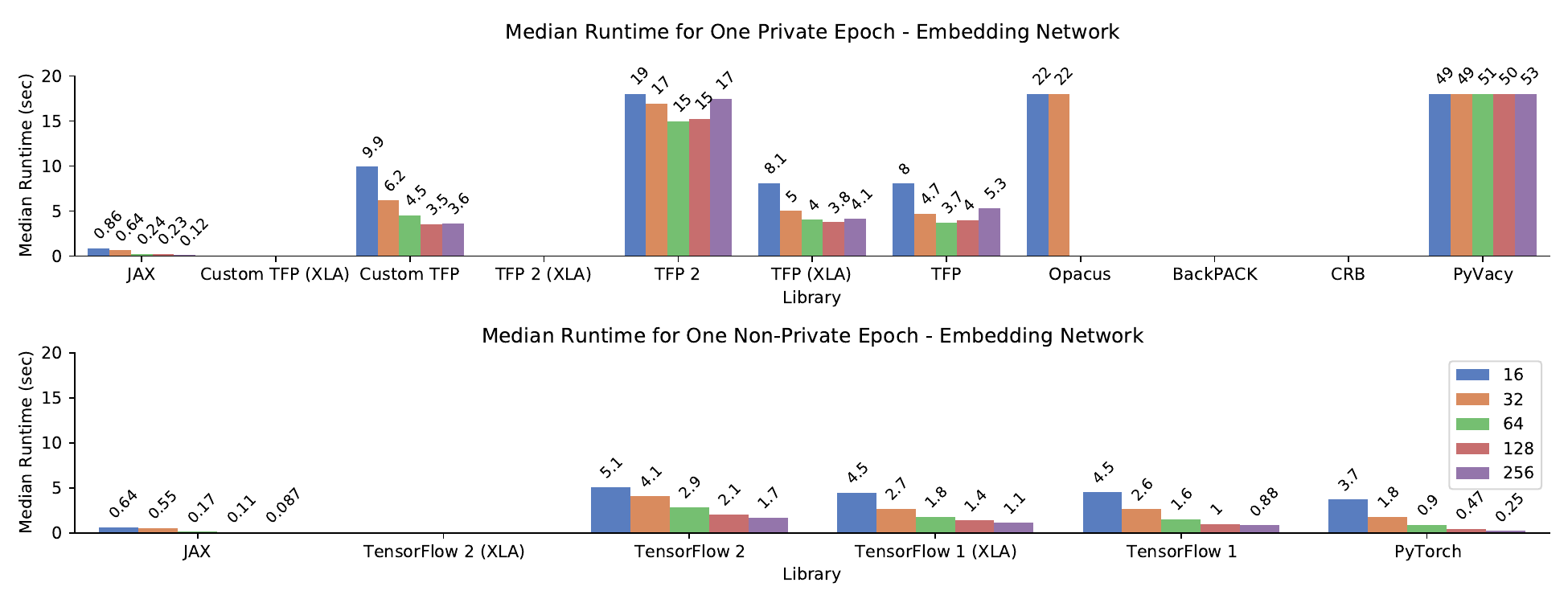}
  \caption{
    \textbf{Runtimes for the Embedding network on the IMDb dataset.}
    JAX is an order of magnitude faster than the best alternatives in the private setting.
    The quadratic memory cost of Opacus prevents evaluation at batch sizes larger than 32.
    An open TensorFlow 2 bug prevents us from evaluating \ctfp (XLA) in this setting~\cite{embedxla}.
    BackPACK and CRB do not support embedding layers.
    The y-axis is truncated for clarity.
  }
  \label{fig:embedplot}
\end{figure*}

Next, we evaluate the embedding network (Figure~\ref{fig:embedplot}), where we see a more exaggerated difference between JAX and the other frameworks.
A (different and also confirmed) TensorFlow 2 bug~\cite{embedxla} and lack of support from BackPACK and CRB prevent us from evaluating this setting on those implementations.
We see a similar phenomenon to the MNIST CNN case for TFP: at larger batch sizes, the library fails to parallelize effectively, thus pessimizing the runtime.

\begin{figure*}[ht!]
  \centering
  \includegraphics[width=16cm]{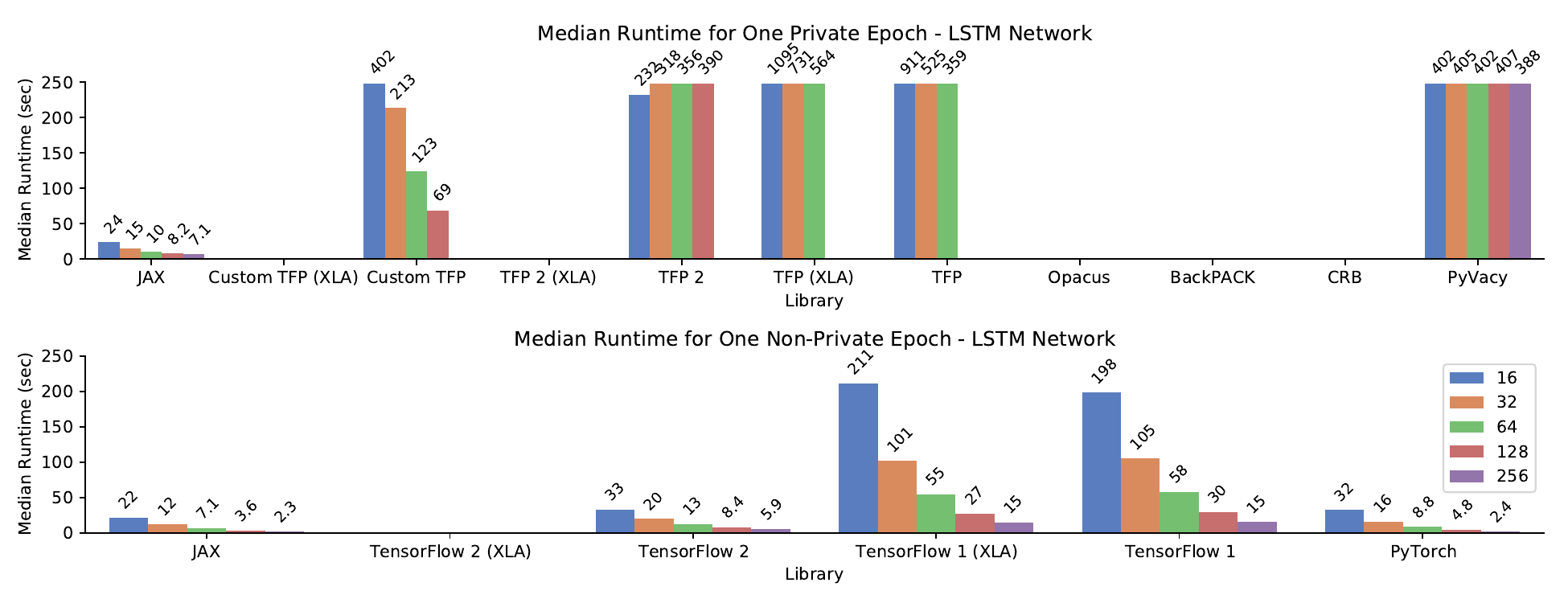}
  \caption{
    \textbf{Runtimes for the LSTM network on the IMDb dataset.}
    JAX is by far the fastest option, resulting in a roughly 50x speedup for batch size 256.
    The quadratic memory cost of Opacus prevents us from evaluating this implementation at these batch sizes, however, we observe a median runtime of 1024.16s at batch size 10.
    Excessive memory consumption prevent us from evaluating \ctfp and TFP at larger batch sizes.
    An open TensorFlow 2 bug prevents us from evaluating \ctfp (XLA) in this setting~\cite{embedxla}.
    BackPACK and CRB do not support embedding layers.
    The y-axis is truncated for clarity.
  }
  \label{fig:lstmplot}
\end{figure*}

For the final runtime experiment we evaluate the LSTM network (Figure~\ref{fig:lstmplot}) and observe similar trends to the embedding network.
Here, however, TensorFlow and PyTorch benefit from a fast cuDNN LSTM implementation in the non-private case which they fail to leverage in the private case, explaining the significant difference in performance.
JAX, on the other hand, uses an LSTM implementation based on primitive operations, which allows it to retain similar performance in both the private and non-private settings.

\begin{table}[ht!]
  \centering
  \begin{tabular}{ccrrrrr}
    \toprule
    \vmap     & \jit      &   Logistic &    FCNN   & MNIST CNN & CIFAR10 CNN & Embedding \\ 
    \midrule
              &           &       1040 &      1240 &      1530 &   3840      & 856 \\ 
    \ding{51} &           &       16.4 &      19.2 &      24.6 &   64.8        & 14.4 \\ 
              & \ding{51} &       1.08 &      2.85 &      22.3 &   84.0       & 3.80 \\ 
    \ding{51} & \ding{51} & \bf{0.231} & \bf{0.239} & \bf{0.535} &  \bf{7.28}      & \bf{0.231} \\ 
    \bottomrule
  \end{tabular}
  \caption{
  \textbf{Ablation of JIT compilation and vectorization in JAX.}
  Median runtime per epoch (seconds) for a run of 20 epochs at batch size 128 for DPSGD in JAX.
  While JIT provides the largest boost in most cases, VMAP also provides a significant performance improvement.
  The two together provide the largest improvement.
  We exclude LSTMs, as JAX runs out of memory without JIT compilation for this model. 
  }
  \label{tab:jaxablation}
\end{table}
To understand the importance of vectorization (via \vmap) and JIT compilation (\jit), we ablate JAX's performance on these tasks with and without these two components (Table~\ref{tab:jaxablation}).
We observe that \jit alone provides up to a 963x improvement, and \vmap alone provides up to a 64x improvement.
While in most cases \jit provides the larger improvement, on the CIFAR10 CNN we observe that \vmap provides a greater reduction.
This is because the model is large enough to sufficiently utilize the GPU with its convolutions and hide the Python overhead through asynchronous execution.
When used in tandem both complement each other, providing a 5160x improvement.

\begin{table}[ht!]
  \centering
  \begin{tabular}{lrrrrrr}
    \toprule
    Mode                  & Logistic & FCNN &   MNIST CNN & CIFAR10 CNN & Embedding & LSTM \\
    \midrule
    Eager                 & 86.3       & 147        & 303       & 561        & 163       & 717 \\
    Graph\                & 9.93       & 12.7       & 32.5      & 85.5       & 22.2      & 361 \\
    XLA                   & 0.76       & 1.83       & 19.5      & 71.5       &           &   \\
    Graph+Vectorization   & \bf{0.703} & 0.770      & 9.07      & 33.8       & \bf{3.53} & 68.6 \\
    XLA+Vectorization     &            & \bf{0.209} & \bf{6.28} & \bf{23.3}  &           & \\
    \bottomrule
  \end{tabular}
  \caption{
  \textbf{Ablation of JIT compilation and vectorization in \ctfp.}
  Median runtime per epoch for a run of 20 epochs at batch size 128 for DPSGD in \ctfp.
  Eager uses no compilation or vectorization.
  Graph refers to TensorFlow's normal graph mode compilation, while XLA represents JIT compilation through XLA.
  In TensorFlow, one can not use vectorization without graph compilation, which is why there is no standalone vectorization.
  Empty entries are due to confirmed active bugs in TensorFlow~\cite{logregxla, embedxla}.
  }
  \label{tab:tfablation}
\end{table}

Similarly, we ablate the components in \ctfp (Table~\ref{tab:tfablation}).
In TensorFlow, \texttt{vectorized\_map} automatically compiles the code, preventing us from ablating it alone.
We observe that XLA in TensorFlow 2 is able to reclaim performance lost by not vectorizing, seeing that the non-VMAP XLA performance comes close to VMAP Graph performance in many settings.
We further see the non-compiled runtimes are not as extreme as seen in JAX: TensorFlow is better optimized to run reasonably fast in all settings.

\begin{table}[ht!]
  \centering
  \begin{tabular}{lrrr}
    \toprule
    Library            & MNIST CNN      & CIFAR10 CNN & IMDb LSTM \\
    \midrule
    JAX                &   187,136      & 10,448      &  \bf{11,984}  \\
    TensorFlow 2 (XLA) &   \bf{271,104} & \bf{15,040} &          \\
    TensorFlow 2       &   179,328      & 11,328      &   9,221  \\
    TensorFlow 1 (XLA) &   157,312      & 10,880      &   5,070  \\
    TensorFlow 1       &   186,368      & 11,480      &   5,264  \\
    PyTorch            &   113,664      & 10,752      &   9,943  \\
    \midrule
    JAX (DP)           &   116,480      & \bf{4,264}  &   \bf{2,487}  \\
    \ctfp (XLA)        &   \bf{137,856} & 3,144       &          \\
    \ctfp              &    57,216      & 1,944       &     137  \\
    TFP 2 (XLA)        &       560      & 168         &        \\
    TFP 2              &       412      & 104         &     105  \\
    TFP (XLA)          &       560      & 168         &      88  \\
    TFP                &    36,608      & 104         &     105  \\
    Opacus             &    36,608      & 1,920       &      10  \\
    BackPACK           &    34,048      & 1,216       &          \\
    CRB                &    40,192      & 2,184  &          \\
    PyVacy             &  $\infty$      & $\infty$    & $\infty$ \\
    \bottomrule
  \end{tabular}
  \caption{
  \textbf{Maximum batch size supported by each library before encountering out of memory errors.}
  The top shows non-private training, while the bottom shows DPSGD.
  Hardware was a NVIDIA Titan V with 12 GB of VRAM.
  Missing entries represent missing functionality or bugs in the frameworks. 
  PyVacy handles examples sequentially one at a time regardless of batch size, giving constant memory consumption with respect to batch size.
  We observe XLA-compiled frameworks (JAX, TensorFlow 2, TensorFlow 1) have superior memory management, except TensorFlow 1 in the MNIST CNN case, which we believe to be due to the limited capabilities of the XLA compiler in TensorFlow 1.
  }
  \label{tab:memory}
\end{table}

Finally, we explore the memory consumption behaviour of these implementations (Table~\ref{tab:memory}), observing that running time has a strong negative correlation to memory consumption.
The modern XLA-compiled libraries provide impressive batch size capability: JAX in the private setting for the MNIST CNN can achieve a larger maximum batch size than PyTorch in the non-private case!
Also, all the frameworks except JAX and PyVacy struggle with batch sizes on the LSTM.
Since PyVacy processes examples sequentially, it has a constant memory consumption with respect to batch size, effectively trading off running time for optimal memory use.
The other frameworks are crippled without access to their fused cuDNN LSTM implementation, while JAX has no issues as its LSTM is composed of primitives.
Finally, due to the specialized per-example gradient computation afforded by CRB for convolutions, it shows the best memory utilization among the PyTorch frameworks, even beating \ctfp (without XLA) in the CIFAR10 CNN case.

\section{Discussion}
\label{sec:discussion}
JAX and \ctfp's runtime advancements can be primarily attributed to the advancement of the compiler present in both of these languages. The XLA compiler performs a variety of operations ranging from memory scheduling to kernel fusion. The memory optimizations are vital for larger models where DPSGD becomes a memory-bound algorithm. One of the core features of XLA is buffer reutilization which has a significant impact on the maximum memory used~\cite{XLA}. Furthermore, the memory scheduler can mitigate peak memory usage to prevent a runtime exception for overusing available memory.

The effectiveness of XLA is demonstrated through the peak batch size experiment (which serves as a proxy for memory efficiency): in both the private and non-private settings, XLA far exceeds alternatives in the peak batch size it supports.
Through this experiment, we also see the benefits of using small operation primitives as opposed to large fused kernels: TensorFlow and PyTorch both leverage the optimized cuDNN kernel in the non-private setting for performance~\cite{tfcudnnlstm, ptcudnnlstm}, but cannot in the private setting, leading to significantly worse performance.
Modifying all existing cuDNN kernels to enable use in the private case would require a non-trivial engineering investment.
JAX instead focuses on optimizing operation primitives, so even in foreign computational circumstances, its performance is comparatively strong. Succinctly, JAX sacrifices the ability to use highly optimized fused kernels for generalizability.

We also observe TFP's significant performance deterioration with XLA in this experiment. Due to the nascency of the XLA compiler, it is more deeply integrated into the newer TensorFlow 2 than TensorFlow 1. Thus, the compiler does not as effectively optimize TensorFlow 1 code, and can sometimes lead to pessimization of performance. The primary documented and observed optimization is autoclustering of operations, which empirically results in suboptimal performance. This carries through to the remaining experiments as well: although we were able to use TensorFlow 1 with XLA, its performance is nowhere near as strong as JAX or TensorFlow 2.

Our implementation of \ctfp leverages the vectorized map for both the forward and backward pass, unlike existing implementations which only use vectorized map for the backward pass (identical to the JAX version).
This gives the compiler explicit information about the independence of batches in the computation, enabling significant optimizations, explaining the improvement in \ctfp compared to the existing TensorFlow implementations.
Details about the implementation are provided in the supplementary code.

We notice that the runtimes are different between \ctfp and JAX despite having the same backend compiler. The primary reason for this is in difference in translation and compilation procedures between the two of them. For example, JAX has a feature called \emph{omnistaging}~\cite{jaxomnistaging} which allows more computation to be visible to the compiler, enabling further optimizations. We did not find evidence or documentation for similar functionality in TensorFlow.

We also observe that simple code segments get translated to different XLA code. Moreover, these differences compound for the larger models dealt with in Section~\ref{sec:experiments}.
In Section~\ref{sec:xla}, we present a running example to demonstrate the divergence between JAX and TensorFlow's XLA assembly (which is a representation of the XLA-compiled module). The difference begins to arise with a call to the gradient function in both frameworks. In our toy example, JAX is able to compile the entire computation graph into a single fused kernel, while TensorFlow compiles the graph into two kernels. In this case, we observe that TensorFlow is slower than JAX (albeit not by a significant margin). Similarly, there are larger differences in the XLA logs of the models discussed in Section \ref{sec:experiments} and these likely contribute to the different performance that is seen across JAX and TensorFlow notwithstanding the same backend.

However, it is not true by default that JAX's compiled code runs faster than TensorFlow 2's. A counter-example to this is seen in Figure~\ref{fig:ffnnplot} where we notice that JAX is slower than \ctfp in the private setting. Thus, while evident that XLA and vectorization present tremendous speedups without loss of generality, their behaviour is rather different depending on the frontend.


Through the ablation study of \vmap and \jit in JAX shown in Table~\ref{tab:jaxablation}, we observe that both components complement each other and enable the level of performance we see.
In general, we observe that JIT provides the larger performance gain, as shown with the FCNN and Embedding result.
For the CNN, the large matrix operations coupled with JAX's asynchronous execution~\cite{jaxasync} allow reasonable utilization of the GPU even without \vmap, which is why we observe less of an improvement from \jit alone in this experiment.
The runtimes without these two primitives is significantly slower than the other frameworks---this is because JAX was built ground-up to leverage these primitives~\cite{BradburyFHJLMW18}.

In the ablation for \ctfp in Table~\ref{tab:tfablation}, we see some key differences with JAX.
First, the mechanism for \vmap in TensorFlow 2 is different from that in JAX: JAX performs op-by-op batching without compilation, while TensorFlow does not~\cite{tfvmap}.
We observe that the non-compiled code still runs in a reasonable amount of time since TensorFlow is optimized to have a competitive eager execution when compared to PyTorch, while JAX is not.
Also, in TensorFlow, XLA's JIT compilation without vectorization is often able to bring the runtime performance close to that of the graph mode with vectorization, implying that XLA is able to recognize and implement some parallelization even when the user does not explicitly request it.
Finally, we observe the benefit of having a fast, fused implementation for LSTMs: while JAX ran out of memory outside of a compiled and vectorized context, \ctfp is able to achieve reasonable runtimes by leveraging the fused kernel.

In the process of conducting our experiments, we observed a new release of JAX (version 0.1.75 to 0.2.0) during which we saw better compilation times and improved running times. This is a major advantage of aligning the differentially private machine learning community with language primitives in JAX and \ctfp. Researchers and engineers alike can take advantage of advancements made to the language core. Additionally, the generalizability of JAX and \ctfp are vital for enabling DPSGD across a variety of domains, and the inability of other frameworks to do so stagger the development of private machine learning.

\subsection{Recommendations}
\label{sec:recs}
JAX and TensorFlow present language primitives that are absent or upcoming in PyTorch~\cite{pytorchvmap}. We encourage language developers to incorporate \vmap and utilize XLA as a compilation path to yield the best runtimes for DPSGD. While these constructs may not be the only way to enable fast DPSGD, given the relevancy of machine learning frameworks, they seem to be the shortest path to fast DPSGD across in the context of machine learning.


In terms of developing future differential privacy libraries and frameworks, there are several key takeaways.
First, leveraging small, generalizable operations and compiler infrastructure are more reliable than hand tuned kernels.
Especially in an emerging field like differential privacy, it is impossible to anticipate how a library or framework will be used, so optimizing for generality will aid in new research.
We observe this in our setting, particularly with the LSTM: as soon as TensorFlow was unable to use the highly optimized cuDNN kernel, its performance dropped significantly, unlike JAX.
Second, maintaining a small, precise API surface will produce a more robust system, as well as one more amenable to optimization.
JAX's small API combined with its strict functional style allow its developers to more easily produce a library that is often more performant than mature counterparts.
Larger APIs such as TensorFlow, while convenient for users, makes it much more prone to bugs and much harder to optimize, as evidenced by the bugs we faced with \ctfp with XLA.

In industry, engineers must port models to production environments and in this setting, \ctfp is an ideal choice.
TensorFlow possesses a custom library for porting models to production, TensorFlow Extended (TFX)~\cite{tfx}.
As a result of the runtime reduction, engineers would be able to prototype and transport models from development environments to production environments while staying within the same framework.
This mitigates potential bugs that could arise from differences in development and production.
JAX is slowly making progress in this area with the experimental \texttt{jax2tf} converter~\cite{jax2tf}, which allows developers to convert a JAX function into a function which uses only TensorFlow operations, allowing deployment of JAX-developed models using TensorFlow infrastructure.

\subsection{Drawbacks}
While integrating XLA into DPSGD presents a massive runtime and memory advantage, there is a cost.
XLA is a subset of all permissable operations in JAX and TensorFlow.
As a result, a researcher or engineer has to be cognizant of the permissible operations to prevent possible errors when incorrectly using XLA.
For example, if dynamic behavior such as \texttt{np.unique(x)} is enabled within an XLA compiled function, an exception is raised.
More subtle errors involving unintended recompilation of code are also possible which can lead to enormous slowdowns.

Now we consider some of the criticisms that are specific to JAX. JAX's primary programming model is functional which diverges from PyTorch and TensorFlow's existing APIs. While this may be an advantage for users experienced in functional programming, there would be a cost to port a user over since it requires understanding the JAX programming paradigm. Moreover, JAX is also in version 0.2.0 compared to PyTorch's 1.6.0 and TensorFlow's 2.4.0 and as a result, its ecosystem is not as mature as its counterparts.
This can present a roadblock to users of who rely on libraries built on top of PyTorch/TensorFlow for research and production.

\section{Conclusion and Future Work}
We have demonstrated that language primitives like vectorization, JIT compilation, and static graph optimization can dramatically improve the running time of private machine learning, realized by JAX and our \ctfp.
In particular, we find that using JAX can almost entirely remove the computational overhead introduced by DPSGD, thus alleviating a major pain point of private machine learning practitioners.

In our work, we focus on conventional set-ups for academic researchers; for future work, it would be insightful to explore the performance of distributed DPSGD, as distributed set-ups are becoming increasingly commonplace.
Furthermore, implementing a PyTorch \jit compatible version of DPSGD could provide an alternative to TensorFlow and JAX, particularly if said implementation is compatible with PyTorch XLA.
Though these two compilation systems are immature compared to TensorFlow and JAX, they are rapidly improving and should not be ignored.
Apart from the facilities available in Python, there are powerful autodifferentiation methods in other more perfomant languages such as Julia~\cite{juliadiff} and Swift~\cite{swiftdiff} which are worthy of study in the context of DPSGD.

\anon{
\section*{Acknowledgments}
We would like to thank Roy Frostig for helpful discussions on JAX, Steve Chien and Shuang Song for their work in implementing our improvements in TensorFlow Privacy, Xi He and Om Thakkar for valuable feedback on drafts of this work, and several JAX, TensorFlow, and Opacus developers who helped answer our issues, including James Bradbury, Peter Buchlovsky, Peter Hawkins, Matthew Johnson, Karthik Prasad, Github handle ravikyram, and Qianli Scott Zhu.
}

\bibliographystyle{alpha}
\bibliography{biblio}
\pagebreak
\appendix

\section{XLA Log Comparison}
\label{sec:xla}

We first construct the simplest component to a neural network, an affine transformation (Figure~\ref{code:pythonsimple}), and observe the XLA logs post-optimization for them. The logs for the JAX and TensorFlow code are in Figures~\ref{code:jaxxla} and~\ref{code:tfxla}, respectively. 
Note that Figure~\ref{code:jaxxla} and Figure~\ref{code:tfxla} are both text files, so they are not indented.

\begin{figure*}[h!]
\begin{verbatim}
    import jax.numpy as jnp
    import jax
    import numpy as np
    import tensorflow as tf

    W = np.random.randn(5, 5).astype(np.float32)  # 5 x 5 Weight Matrix
    b = np.random.randn(5).astype(np.float32)  # 5 x 1 bias vector
    x = np.random.randn(5).astype(np.float32)  # input x

    @jax.jit  # enables XLA + JIT
    def matvec(W, x, b):
        return jnp.dot(W, x) + b  # W * x + b

    @tf.function(experimental_compile=True)  # enables XLA + JIT
    def matvec2(W, x, b):
        return tf.linalg.matvec(W, x) + b
\end{verbatim}
  \caption{Matrix-Vector Product in JAX and TensorFlow}
\label{code:pythonsimple}
\end{figure*}

\begin{figure*}[h!]
\begin{verbatim}
    HloModule jit_fn1.8

    %scalar_add_computation (scalar_lhs: f32[], scalar_rhs: f32[]) -> f32[] {
      %scalar_lhs = f32[] parameter(0)
      %scalar_rhs = f32[] parameter(1)
      ROOT %add = f32[] add(f32[] %scalar_lhs, f32[] %scalar_rhs)
    }

    %fused_computation (param_0.1: f32[5], param_1.2: f32[5,5],
     param_2.2: f32[5]) -> f32[5] {
      %param_1.2 = f32[5,5]{1,0} parameter(1)
      %param_2.2 = f32[5]{0} parameter(2)
      %broadcast.1 = f32[5,5]{1,0} broadcast(f32[5]{0} %param_2.2),
       dimensions={1}
      %multiply.1 = f32[5,5]{1,0} multiply(f32[5,5]{1,0} %param_1.2,
      f32[5,5]{1,0} %broadcast.1)
      %constant_1 = f32[] constant(0)
      %reduce.1 = f32[5]{0} reduce(f32[5,5]{1,0} %multiply.1, f32[] %constant_1),
                dimensions={1}, to_apply=%scalar_add_computation,
                metadata={op_type="dot_general" op_name="jit(fn1)/

    dot_general[ dimension_numbers=(((1,), (0,)), ((), ()))
    \n precision=None ]"
                 source_file="<ipython-input-3-8e7c515249a1>"
                  source_line=3}
      %param_0.1 = f32[5]{0} parameter(0)
      ROOT %add.1 = f32[5]{0} add(f32[5]{0} %reduce.1,
                    f32[5]{0} %param_0.1),
                    metadata={op_type="add" op_name="jit(fn1)/add"
                     source_file="<ipython-input-3-8e7c515249a1>"
                     source_line=3}
    }

    ENTRY %jit_fn1.8 (parameter.1: f32[5,5], parameter.2: f32[5],
     parameter.3: f32[5]) -> (f32[5]) {
      %parameter.3 = f32[5]{0} parameter(2)
      %parameter.1 = f32[5,5]{1,0} parameter(0)
      %parameter.2 = f32[5]{0} parameter(1)
      %fusion = f32[5]{0} fusion(f32[5]{0} %parameter.3,
      f32[5,5]{1,0} %parameter.1, f32[5]{0} %parameter.2),
       kind=kLoop, calls=%fused_computation,
       metadata={op_type="add" op_name="jit(fn1)/add"
        source_file="<ipython-input-3-8e7c515249a1>" source_line=3}
      ROOT %tuple.7 = (f32[5]{0}) tuple(f32[5]{0} %fusion)
    }
\end{verbatim}
\caption{JAX XLA Logs}
\label{code:jaxxla}
\end{figure*}

\begin{figure*}[h!]
\begin{verbatim}
    HloModule a_inference_loss_nograd_15__XlaMustCompile_true_config_proto___
    n_007_n_003CPU_020_001_n_007_n_003GPU_020_0012_005__0010J_0008_
    001_202_001_000__executor_type____.15

    %scalar_add_computation (scalar_lhs: f32[], scalar_rhs: f32[]) -> f32[] {
      %scalar_lhs = f32[] parameter(0)
      %scalar_rhs = f32[] parameter(1)
      ROOT %add = f32[] add(f32[] %scalar_lhs, f32[] %scalar_rhs)
    }

    %fused_computation (param_0.1: f32[5], param_1.2: f32[5,5],
                        param_2.2: f32[5]) -> f32[5] {
      %param_1.2 = f32[5,5]{1,0} parameter(1)
      %param_2.2 = f32[5]{0} parameter(2)
      %broadcast.1 = f32[5,5]{1,0} broadcast(f32[5]{0} %param_2.2),
       dimensions={1}
      %multiply.1 = f32[5,5]{1,0} multiply(f32[5,5]{1,0} %param_1.2,
                    f32[5,5]{1,0}
                    %broadcast.1)
      %constant_1 = f32[] constant(0)
      %reduce.1 = f32[5]{0} reduce(f32[5,5]{1,0} %multiply.1, f32[] %constant_1),
                dimensions={1}, to_apply=%scalar_add_computation,
                 metadata={op_type="Squeeze" op_name="MatVec/Squeeze"}
      %param_0.1 = f32[5]{0} parameter(0)
      ROOT %add.1 = f32[5]{0} add(f32[5]{0} %reduce.1, f32[5]{0} %param_0.1),
                    metadata={op_type="AddV2" op_name="add"}
    }

    ENTRY %a_inference_loss_nograd_15__XlaMustCompile_true_config_proto___
    n_007_n_003CPU_020_001_n_007_n_003GPU_020_0012_005__0010J_0008_001_
    202_001_000__executor_type____.15
    (arg0.1: f32[5,5], arg1.2: f32[5], arg2.3: f32[5]) -> f32[5] {
      %arg2.3 = f32[5]{0} parameter(2), parameter_replication={false},
      metadata={op_name="XLA_Args"}
      %arg0.1 = f32[5,5]{1,0} parameter(0), parameter_replication={false},
       metadata={op_name="XLA_Args"}
      %arg1.2 = f32[5]{0} parameter(1), parameter_replication={false},
       metadata={op_name="XLA_Args"}
      ROOT %fusion = f32[5]{0} fusion(f32[5]{0} %arg2.3, f32[5,5]{1,0} %arg0.1,
                    f32[5]{0} %arg1.2), kind=kLoop, calls=%fused_computation,
                    metadata={op_type="AddV2" op_name="add"}
    }
\end{verbatim}
\caption{The XLA logs for the TensorFlow function}
\label{code:tfxla}
\end{figure*}

\clearpage
Figure~\ref{code:jaxxla} and Figure~\ref{code:tfxla} both possess a single fused kernel performing the same operation. 
We observe barring language-specific XLA syntax, the computation graph generated in the XLA log files are the same. 
This implies that the computation path taken by JAX and TensorFlow in this case are equivalent.
Now, in Figure~\ref{code:advanced}, we extend the code in Figure~\ref{code:pythonsimple} to also take the gradients with respect to $W$ and $b$, which are the parameters of the network. 
We now modify the example as follows:

\begin{figure*}[h!]
    \begin{verbatim}
        @jax.jit
        @jax.grad
        def grad_loss_jax(W, b, x targ):
            out = jnp.dot(W, x) + b
            return jnp.sum((out - targ)**2)

        @tf.function(experimental_compile=True)
        def grad_loss_tf(W, b, x, targ):
            with tf.GradientTape() as tape:
                tape.watch(W)
                out = tf.linalg.matvec(W, x) + b
                loss = tf.reduce_sum((out - targ)**2)
            return tape.gradient(loss, W)
    \end{verbatim}
  \caption{Matrix-Vector Product with Gradients in JAX and TensorFlow}
  \label{code:advanced}
\end{figure*}

These two code blocks do precisely the same computation once again, however, in this case result in different XLA log files.
See Figures~\ref{code:gradjaxxla} and~\ref{code:gradtfxla}.

\begin{figure*}
    \begin{verbatim}
%fused_computation (param_0.5: f32[5], param_1.10: f32[5],
                    param_2.6: f32[5], param_3.5: f32[5,5]) -> f32[5,5] {
  %constant_16 = f32[] constant(2), metadata={op_type="mul"
   op_name="jit(loss)/jit(jvp(loss))/mul" source_file="<ipython-input-73-db04383b747a>"
    source_line=8}
  %broadcast.6 = f32[5]{0} broadcast(f32[] %constant_16),
   dimensions={}, metadata={op_type="mul"
   op_name="jit(loss)/jit(jvp(loss))/mul" source_file="<ipython-input-73-db04383b747a>"
    source_line=8}
  %param_3.5 = f32[5,5]{1,0} parameter(3)
  %param_2.6 = f32[5]{0} parameter(2)
  %broadcast.10 = f32[5,5]{1,0} broadcast(f32[5]{0} %param_2.6), dimensions={1}
  %multiply.9 = f32[5,5]{1,0} multiply(f32[5,5]{1,0} %param_3.5,
  f32[5,5]{1,0} %broadcast.10)
  %constant_20 = f32[] constant(0)
  %reduce.4 = f32[5]{0} reduce(f32[5,5]{1,0} %multiply.9, f32[] %constant_20), dimensions={1},
   to_apply=%scalar_add_computation,
   metadata={op_type="dot_general"
   op_name="jit(loss)/jit(jvp(loss))/jit(jvp(fn1))/
   dot_general[ dimension_numbers=(((1,), (0,)), ((), ()))  \n precision=None ]"
    source_file="<ipython-input-73-db04383b747a>"  source_line=3}
  %param_1.10 = f32[5]{0} parameter(1)
  %add.3 = f32[5]{0} add(f32[5]{0} %reduce.4,
  f32[5]{0} %param_1.10), metadata={op_type="add"
  op_name="jit(loss)/jit(jvp(loss))/jit(jvp(fn1))/add"
  source_file="<ipython-input-73-db04383b747a>"  source_line=3}
  %param_0.5 = f32[5]{0} parameter(0)
  %subtract.1 = f32[5]{0} subtract(f32[5]{0} %add.3,
   f32[5]{0} %param_0.5), metadata={op_type="sub"
      op_name="jit(loss)/jit(jvp(loss))/sub"
    source_file="<ipython-input-73-db04383b747a>" source_line=8}
  %multiply.6 = f32[5]{0} multiply(f32[5]{0} %broadcast.6,
   f32[5]{0} %subtract.1), metadata={op_type="mul" op_name="jit(loss)/jit(jvp(loss))/mul"
   source_file="<ipython-input-73-db04383b747a>" source_line=8}
  %broadcast.5 = f32[5,5]{1,0} broadcast(f32[5]{0} %multiply.6),
  dimensions={0}
  ROOT %multiply.5 = f32[5,5]{1,0} multiply(f32[5,5]{1,0} %broadcast.5,
  f32[5,5]{1,0} %broadcast.10),  metadata={op_type="dot_general" op_name="jit(loss)/
  jit(transpose(jvp(loss)))/jit(transpose(jvp(fn1)))/dot_general[
       dimension_numbers=(((), ()), ((), ()))\n precision=None ]"
       source_file="<ipython-input-73-db04383b747a>" source_line=3}
}

    \end{verbatim}
    \caption{JAX post-optimization Gradient XLA Kernel. Note that there is only one fused kernel titled \texttt{fused\_computation}. For brevity, the entry point to the XLA computation graph and auxiliary information are omitted.}
\label{code:gradjaxxla}
\end{figure*}

\begin{figure*}[h!]
\begin{verbatim}

    %fused_computation (param_0.5: f32[5], param_1.7: f32[5], param_2.5:
     f32[5], param_3.3: f32[5]) -> f32[5,5] {
      %constant_1 = f32[] constant(2), metadata={op_type="Mul"
      op_name="PartitionedCall_1/gradients/pow_grad/mul_1"}
      %broadcast.4 = f32[5]{0} broadcast(f32[] %constant_1), dimensions={}
      , metadata={op_type="Mul" op_name="PartitionedCall_1/gradients/
      pow_grad/mul_1"}
      %param_1.7 = f32[5]{0} parameter(1)
      %param_2.5 = f32[5]{0} parameter(2)
      %add.1 = f32[5]{0} add(f32[5]{0} %param_1.7, f32[5]{0} %param_2.5),
      metadata={op_type="AddV2" op_name="PartitionedCall/add"}
      %param_0.5 = f32[5]{0} parameter(0)
      %subtract.0 = f32[5]{0} subtract(f32[5]{0} %add.1, f32[5]{0}
      %param_0.5), metadata={op_type="Sub" op_name="PartitionedCall/sub_0"}
      %multiply.3 = f32[5]{0} multiply(f32[5]{0} %broadcast.4, f32[5]{0}
      %subtract.0), metadata={op_type="Mul" op_name="PartitionedCall_1/
      gradients/pow_grad/mul_1"}
      %broadcast.3 = f32[5,5]{1,0} broadcast(f32[5]{0} %multiply.3),
      dimensions={0}
      %param_3.3 = f32[5]{0} parameter(3)
      %broadcast.5 = f32[5,5]{1,0} broadcast(f32[5]{0} %param_3.3),
      dimensions={1}
      ROOT %multiply.2 = f32[5,5]{1,0} multiply(f32[5,5]{1,0} %broadcast.
      3, f32[5,5]{1,0} %broadcast.5), metadata={op_type="MatMul"
      op_name="PartitionedCall_1/gradients/MatVec/MatMul_grad/MatMul"}
    }

    %fused_computation.1 (param_0.4: f32[5,5], param_1.8: f32[5]) -> f32
    [5] {
      %param_0.4 = f32[5,5]{1,0} parameter(0)
      %param_1.8 = f32[5]{0} parameter(1)
      %broadcast.6 = f32[5,5]{1,0} broadcast(f32[5]{0} %param_1.8),
      dimensions={1}
      %multiply.4 = f32[5,5]{1,0} multiply(f32[5,5]{1,0} %param_0.4, f32[5,
      5]{1,0} %broadcast.6)
      %constant_2 = f32[] constant(0)
      ROOT %reduce.1 = f32[5]{0} reduce(f32[5,5]{1,0} %multiply.4, f32[]
      %constant_2), dimensions={1}, to_apply=%scalar_add_computation,
      metadata={op_type="Squeeze" op_name="PartitionedCall/MatVec/Squeeze"}
    }

\end{verbatim}
\caption{TensorFlow post-optimization Gradient XLA Kernel. Note that there are two fused kernels titled \texttt{fused\_computation} and \texttt{fused\_computation.1}. For brevity, the entry point to the XLA computation graph and auxiliary information are omitted.}
\label{code:gradtfxla}
\end{figure*}
\clearpage

Here we notice a considerable difference. 
Observe that the XLA compiler for JAX creates one large fused kernel that runs, as opposed to TensorFlow's multiple kernels that run. 
In this case, JAX is faster than TensorFlow as a consequence of minor overhead involved with two kernel launches compared to one. 
However, one should not expect that multiple kernel launches always implies slower performance. 
In certain cases, with appropriate kernel scheduling, multiple launches can actually be faster, as we observe in our runtime experiments when \ctfp is faster than JAX.

\section{Architectures}
\label{sec:arch}

We describe the precise architectures of the deep networks used for the experiments. 
All the networks are sequential, that is, the output of any layer is fed into only the subsequent layer. 
All networks use the ReLU activation function between linear layers, so we exclude them from our definitions for brevity.
Further documentation of these architectures appears in our Github repository at \url{https://github.com/TheSalon/fast-dpsgd}.

\begin{table}[h!]
  \centering
  \begin{tabular}{lrr}
    \toprule
    Layer Type & Input Features & Output Neurons \\
    \midrule
    Fully Connected & 104 & 50 \\
    Fully Connected & 50 & 10 \\
    \bottomrule
  \end{tabular}
  \caption{\textbf{FFNN Architecture}}
  \label{tab:ffnn_arch}
\end{table}

\begin{table}[h!]
  \centering
  \begin{tabular}{lrrrrr}
    \toprule
    Layer Type     & Input Filters/Features & Output Filters/Neurons & Filter Size & Stride & Padding\\
    \midrule
    2D Convolution  & 1   & 16 & $8\times 8$ & 2 & 3 \\
    2D MaxPool      & 16  & 16 & $2\times 2$ & 1 & 0 \\
    2D Convolution  & 16  & 32 & $4\times 4$ & 1 & 0 \\
    2D MaxPool      & 32  & 32 & $2\times 2$ & 1 & 0 \\
    2D Convolution  & 32  & 32 & $4\times 4$ & 1 & 0 \\
    Fully Connected & 512 & 32 & \\
    Fully Connected &  32 & 10 & \\
    \bottomrule
  \end{tabular}
  \caption{\textbf{MNIST CNN Architecture}}
  \label{tab:mnist_arch}
\end{table}

\begin{table}[h!]
  \centering
  \begin{tabular}{lrrrrr}
    \toprule
    Layer Type     & Input Filters & Output Filters & Filter Size & Stride & Padding\\
    \midrule
    2D Convolution  & 3   & 32 & $3\times 3$ & 1 & 1 \\
    2D Convolution  & 32  & 32 & $3\times 3$ & 1 & 1 \\
    2D Average Pool & 32  & 32 & $2\times 2$ & 2 & 0 \\

    2D Convolution  & 32  & 64 & $3\times 3$ & 1 & 1 \\
    2D Convolution  & 64  & 64 & $3\times 3$ & 1 & 1 \\
    2D Average Pool & 64  & 64 & $2\times 2$ & 2 & 0 \\

    2D Convolution  & 64  & 128 & $3\times 3$ & 1 & 1 \\
    2D Convolution  & 128 & 128 & $3\times 3$ & 1 & 1 \\
    2D Average Pool & 128 & 128 & $2\times 2$ & 2 & 0 \\

    2D Convolution  & 128 & 256 & $3\times 3$ & 1 & 1 \\
    2D Convolution  & 256 &  10 & $3\times 3$ & 1 & 1 \\
    2D Global Average Pool & 10 & 10 & & & \\
    \bottomrule
  \end{tabular}
  \caption{\textbf{CIFAR10 CNN Architecture}. The Global Average Pool reduces all the spatial dimensions so the resulting tensor only has 10 activations (corresponding to the 10 CIFAR10 classes).}
  \label{tab:cifar_arch}
\end{table}

\begin{table}[h!]
  \centering
  \begin{tabular}{lrr}
    \toprule
    Layer Type & Input Features & Output Neurons \\
    \midrule
    Embedding & 256 & 256 $\times$ 16 \\
    1D Average Pool & 256 $\times$ 16 & 16 \\
    Fully Connected & 16 & 2 \\
    \bottomrule
  \end{tabular}
  \caption{\textbf{Embedding Network Architecture}. The embedding layer has a vocabulary size of 10,004.}
  \label{tab:embed_arch}
\end{table}

\begin{table}[h!]
  \centering
  \begin{tabular}{lrr}
    \toprule
    Layer Type & Input Features & Output Neurons \\
    \midrule
    Embedding & 256 & 256 $\times$ 100 \\
    LSTM & 256 $\times$ 100 & 256 $\times$ 100 \\
    1D Average Pool & 256 $\times$ 100 & 100 \\
    Fully Connected & 100 & 2 \\
    \bottomrule
  \end{tabular}
  \caption{\textbf{LSTM Network Architecture} The embedding layer has a vocabulary size of 10,004. Notice for the LSTM, we return the entire sequence of outputs (not just the final activations). We use static unrolling for JAX with \jit, TFP, and \ctfp. Otherwise, the dynamic LSTM implementation is used. In PyTorch and TensorFlow, the dynamic LSTM implementation uses the optimized cudNN kernel (which is not usable in the private case). In JAX, the dynamic LSTM uses the \texttt{lax.scan} primitive, which is slower than unrolling in a compiled context, but faster outside of a compiled context.}
  \label{tab:lstm_arch}
\end{table}

\end{document}